%% file: main.tex
\documentclass[runningheads]{llncs}
\usepackage[T1]{fontenc}
\usepackage{graphicx}

\usepackage{amsmath}
\usepackage{amssymb}
\usepackage{xspace}

\usepackage{marvosym}
\usepackage{ifsym}
\usepackage{algorithmicx,algorithm}

\usepackage{booktabs}
\usepackage{makecell}
\usepackage{diagbox}
\usepackage{multirow}
\usepackage{arydshln}
\usepackage[noend]{algpseudocode}

\usepackage[colorlinks, bookmarks=false,linkcolor=blue, citecolor=blue,urlcolor=blue]{hyperref}

\usepackage{paralist}

\usepackage{color}

\def\onedot{.\xspace}
 
\def\ie{\emph{i.e}\onedot}

\def\etal{\emph{et al}.}

\begin{document}
\title{Propheter: Prophetic Teacher Guided Long-Tailed Distribution Learning
}
%
%
\author{
Wenxiang Xu\inst{1,5}\and
Yongcheng Jing\inst{2}\and
Linyun Zhou\inst{1} \and
Wenqi Huang\inst{3} \and \\
Lechao Cheng\inst{4\textrm{(\Letter)}} \and
Zunlei Feng\inst{1,6} \and
Mingli Song\inst{1,6}
}

\authorrunning{W. Xu et al.}
%
\institute{
Zhejiang University, Hangzhou 310027, China \\
\email{\{xuwx1996, zhoulyaxx, zunleifeng, brooksong\}@zju.edu.cn}
\and
The University of Sydney, Darlington, NSW 2008, Australia \\
\email{yjin9495@uni.sydney.edu.au}
\and
Digital Grid Research Institute, China Southern Power Grid,\\ Guangzhou 510663, China \\
\email{huangwq@csg.cn}
\and
Zhejiang Lab, Hangzhou 311121, China \\
\email{chenglc@zhejianglab.com}
\and
Zhejiang University - China Southern Power Grid Joint Research Centre on AI,\\ Hangzhou 310058, China
\and
ZJU-Bangsun Joint Research Center
}
\maketitle              
\begin{abstract}
The problem of deep long-tailed learning, a prevalent challenge in the realm of generic visual recognition, persists in a multitude of real-world applications. To tackle the heavily-skewed dataset issue in long-tailed classification, prior efforts have sought to augment existing deep models with the elaborate class-balancing strategies, such as class rebalancing, data augmentation, and module improvement. Despite the encouraging performance, the limited class knowledge of the tailed classes in the training dataset still bottlenecks the performance of the existing deep models. In this paper, we propose an innovative long-tailed learning paradigm that breaks the bottleneck by guiding the learning of deep networks with external prior knowledge. This is specifically achieved by devising an elaborated ``prophetic'' teacher, termed as ``Propheter'', that aims to learn the potential class distributions. The target long-tailed prediction model is then optimized under the instruction of the well-trained ``Propheter'', such that the distributions of different classes are as distinguishable as possible from each other. Experiments on eight long-tailed benchmarks across three architectures demonstrate that the proposed prophetic paradigm acts as a promising solution to the challenge of limited class knowledge in long-tailed datasets. The developed code is publicly available at \url{https://github.com/tcmyxc/propheter}.

\end{abstract}
\section{Introduction}
Deep long-tailed learning is a formidable challenge in practical visual recognition tasks.
The goal of long-tailed learning is to train effective models from a vast number of images, but most involving categories contain only a minimal number of samples. Such a long-tailed data distribution is prevalent in various real-world applications, including image classification \cite{Li2022GCL}, object detection \cite{Focalloss}, and segmentation \cite{Ren2020balms}. 
As such, for the minority classes, the lack of sufficient instances to describe the intra-class diversity leads to the challenge of heavily skewed models towards the head classes.

\begin{figure}[!t]
    \centering
    \includegraphics[width=0.65\textwidth]{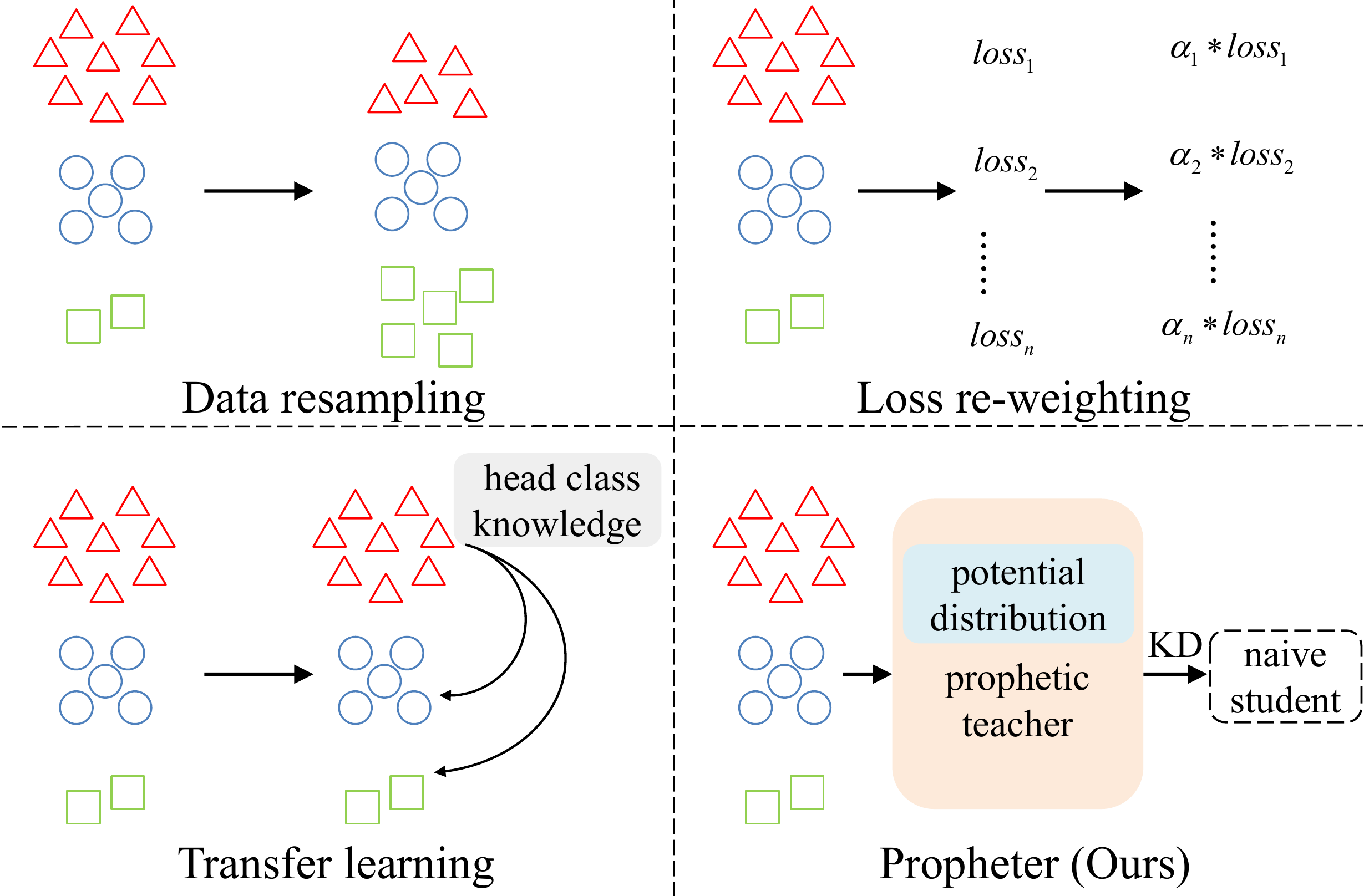}
    \vspace{-2mm}
    \caption{Illustrations of the proposed \emph{Propheter} paradigm as compared with other prevalent long-tailed learning schemes, including data resampling, loss re-weighting, and transfer learning. }
    \label{fig:two_stage}
    \vspace{-4mm}
\end{figure}

To tackle this challenge, various class balancing techniques have been proposed in the community, which can be broadly divided into three categories, \ie, \emph{Data Resampling}, \emph{Loss Re-weighting}, and \emph{Transfer Learning} based methods.
Specifically, \emph{Data Resampling} techniques involve either downsampling the samples in the head class or upsampling those in the tailed class ~\cite{chawla2002smote,feng2021exploring}, whereas the \emph{Loss Re-weighting} methods design the elaborated loss functions to boost the accuracy of unbalanced data, especially those in the tailed classes ~\cite{cui2019CBLoss,Li2022GCL}.
In contrast, the transfer-learning-based approaches enhance the performance of the tailed classes by transferring the knowledge either from the head classes~\cite{jamal2020rethinking} to the tailed ones or from a teacher to a student through knowledge distillation~\cite{iscen2021cbd}.

Despite the plausible performance, most of the existing approaches heavily rely on well-designed sophisticated data samplers to alleviate the dilemma of the scarce tail classes, with some of them even retaining a large number of auxiliary parameters. 
As such, even at the expense of the heavy computational burden, the challenge of limited tailed classes is not fundamentally addressed.
Besides, there is a lack of explicit considerations in the differences of the activation distributions among various long and tail categories in the high-dimensional feature space.

In this paper, we strive to make one further step towards breaking the performance bottleneck of long-tailed representation learning, by devising a novel \emph{``Propheter''} paradigm that explores the long-tailed problem from the perspective of the \emph{latent space}.
Unlike the existing three prevalent schemes illustrated in Fig.~\ref{fig:two_stage}, including data resampling, loss re-weighting, and transfer learning, the proposed \emph{``Propheter''} aims to explicitly learn the potential feature distributions of various long and tail categories, based on which the challenge of deficient tailed data is solved by looking into the learned latent space.  

To achieve this goal, we start by proposing a reasonable assumption that the activation distributions of various categories should be distinguishable in the latent space.
Based on this hypothesis, we develop a residual distribution learning scheme to bridge the gap between the learned and the real class distributions.
Specifically, we develop a parameter-efficient plug-and-play propheter module that adaptively learns the residual between the real and the present head/tailed class distributions in the dataset, by incorporating the captured class-specific distribution knowledge in the devised propheter module.
We then transfer the learned class-specific distribution knowledge from the elaborated propheter module to the vanilla student long-tailed model, through the well-studied knowledge distillation scheme, leading to an innovative prophetic teacher guided long-tailed distribution learning paradigm.

In aggregate, the key contributions of this work can be summarized as follows:
\begin{compactitem}
    \item We propose the conjecture that the performance bottleneck of the existing deep long-tailed representation learning lies in the lack of the explicit considerations in the potential category-specific activation distributions, and accordingly devise a novel \emph{``Propheter''} paradigm that aims to learn the potential real feature distribution of each head/tail class in the dataset;
    \item We propose a dedicated two-phase prophetic training strategy for the proposed \emph{``Propheter''} paradigm, comprising the prophetic teacher learning and the propheter-guided long-tailed prediction stages. 
    The first phase is backed by residual distribution learning, whereas the following propheter-instructed learning stage is built upon knowledge distillation;
    \item Experiments on eight prevalent long-tailed datasets demonstrate that our propheter-guided paradigm delivers encouraging results, with a performance gain of over 4\% on CIFAR-10-LT as compared with the existing methods.
\end{compactitem}

\section{Related Work}
\noindent
\textbf{Long-Tailed Image Categorization.} 
In recent years, the field of computer vision has made remarkable strides
across various domains, such as depth estimation \cite{zhao2020collaborative,zhao2022jperceiver},  intelligent transportation \cite{xi2022data,xi2023incentive,xi2023modeling}, image recognition~\cite{su2021prioritized,su2021locally,su2021bcnet,su2022vitas}, image detection \cite{xi2018detection,zhai2022exploring,zhai2022one}, 3D object analysis \cite{feng2023exploring,jing2023deep,zhao2023bevsimdet},
model reusing \cite{yang2022factorizing,yang2022deep},
dataset condensation~\cite{liu2022dataset,liu2023slimmable,yu2023dataset},
facial expression recognition~\cite{luo2023learning,zhu2023knowledge},  and generative models~\cite{feng2017graph,yang2023diffusion}. 
Among these tasks, the challenge of addressing long-tailed visual recognition has garnered significant attention as a trending topic.
Several approaches~\cite{chawla2002smote,chu2020feature,kang2019decoupling,Focalloss} have been proposed to tackle the class imbalance issue, including decoupling representation and classifier learning. For example, Kang \etal~\cite{kang2019decoupling} employ multiple normalization techniques on the linear classifier layer. Chu \etal~\cite{chu2020feature} enhance the feature representation of the tail classes. Another direction focuses on promoting the learning of the tail classes, such as re-sampling techniques~\cite{chawla2002smote}, loss re-weighting~\cite{cui2019CBLoss}, loss balancing~\cite{Focalloss}, 
and knowledge transfer from the head classes to the tail classes~\cite{jamal2020rethinking}.

\noindent\textbf{Knowledge Distillation.} The concept of knowledge distillation, as first introduced in~\cite{Hinton2015distil}, pertains to the transfer of information from a teacher model to a student model. This technique has been widely employed across various deep learning and computer vision tasks \cite{Hinton2015distil,hou2019learning,iscen2021cbd}.
Iscen \etal~\cite{iscen2021cbd} leveraged \emph{Class-Balanced Distillation} (CBD) to enhance feature representations. This method trains multiple teacher models through different seed or data augmentation techniques. Feature distillation has also been demonstrated to be effective in tasks such as asymmetric metric learning and reducing catastrophic forgetting in incremental learning~\cite{hou2019learning}.

\section{Proposed Method}

\begin{figure}[t]
    \centering
    \includegraphics[width=0.8\textwidth]{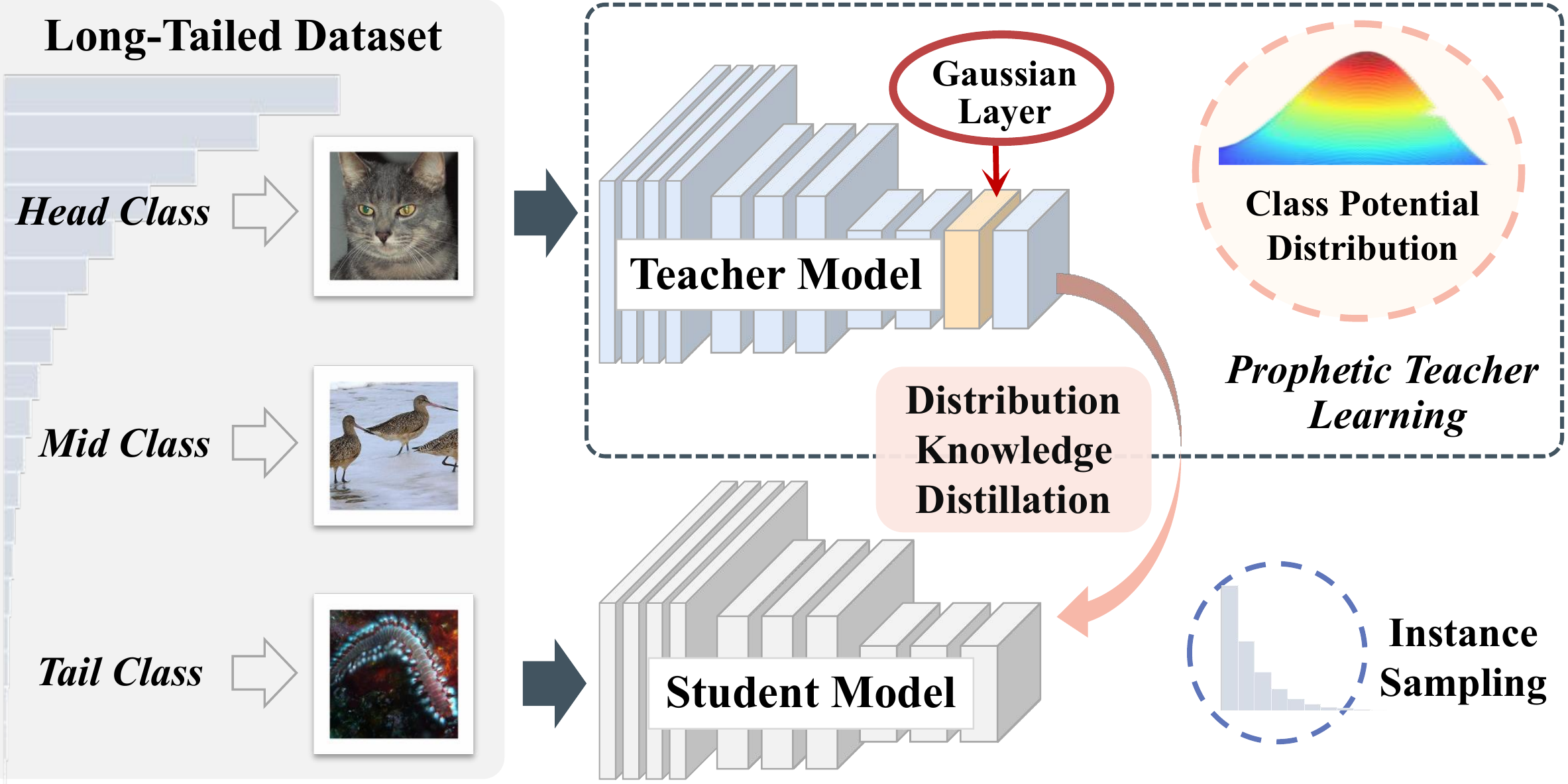}
    \vspace{-2mm}
    \caption{The proposed \emph{{Propheter}-Guided Long-Tailed Learning}.}
    \label{fig:method}
    \vspace{-2mm}
\end{figure}

\subsection{Prophetic Teacher Learning}

\noindent\textbf{Learning Potential Class Distributions.}
In the real world, data often exhibit a specific distribution, such as Gaussian, uniform, or chi-square distributions. When such data are fed into a deep learning model, the features generated by the embedding layer should similarly reflect this distribution. Conventional deep learning models only learn the categorical distribution of the training dataset. However, while commonly used datasets are often balanced, real-world data often exhibit imbalanced distributions. Consequently, deep learning models trained on balanced datasets may not perform well on skewed data.
Hence, it has been observed that relying solely on the original distribution of the training data results in limited generalization ability for deep learning models. By incorporating prior knowledge of the potential distribution of the dataset into the training process of the deep learning model, its performance can be improved as demonstrated in recent works such as \cite{Li2022GCL,Ren2020balms}.

\noindent\textbf{Balanced Propheter Learning.}
While datasets may vary in their distributions across categories, features within a single category are generally assumed to follow a Gaussian distribution~\cite{Li2022GCL}.

Suppose the features of different class samples conform to a Gaussian distribution.
We can derive the feature representation $\mathbf{f}^{pred}_{i}$ for class $i$ from the embedding layer (convolution layer).
However, there exists a disparity between the learned features distribution and the ground-truth distribution $\mathbf{f}^{gt}_{i}$. Given that Gaussian distribution is additive, we can augment the learned Gaussian distribution $\mathbf{E}^{learned}_{i}$ after the convolution layer with the aim of:

\begin{equation}\label{eq1}
\mathbf{f}^{pred}_{i} + \mathbf{E}^{learned}_{i} \triangleq \mathbf{f}^{gt}_{i},
\end{equation}
where $\mathbf{E} \sim \mathcal{N}(\mathbf{u}, \boldsymbol{\sigma^{2}})$. 
From the formula, it can be seen that our goal is to learn the residual between two distributions $\mathbf{E}^{learned}_{i}$. 
Specifically, we will learn an independent Gaussian distribution $\mathbf{E}^{learned}_{i}$ for each class.
For simplicity, the following article we will replace $\mathbf{E}^{learned}_{i}$ with $\mathbf{E}_{i}$.

About the Gaussian distribution, we will sample from the standard normal distribution
$\mathbf{E} \sim \mathcal{N}(0, 1)$. Then we scale and translate the distribution through two learnable parameters $a_{i}$ and $b_{i}$. So the $\mathbf{E}_{i}$ will be:
\begin{equation}\label{eq2}
\mathbf{E}_{i} := a_{i} \cdot \mathbf{E} + b_{i},
\end{equation}
and then $\mathbf{E}_{i} \sim \mathcal{N}(b_{i}, (a_{i})^{2})$. It can be seen that we can change the distribution through these two learnable parameters. That is to say, these two parameters control the shape of the class distribution in the latent space.

However, we discovered that a straightforward implementation of this approach did not yield the desired results. In response, we adopt an alternating training scheme, inspired by Liang \etal~\cite{liang2020training}, to incrementally incorporate the target Gaussian noise into the model. Specifically, as outlined in Alg.~\ref{alg:one_stage_train}, during epochs where $\text{epoch} \ \% \ \text{period} = 0$, we inject Gaussian noise into the features generated by the convolution layer, producing a modified feature vector which is then supplied to the classifier. This pathway is referred to as the Gaussian Noise (GN) path. Conversely, during non-zero values of $\text{epoch} \ \% \ \text{period}$, the standard (STD) path, which does not alter the feature representation, is employed.

\begin{algorithm}[t]
\caption{Training a prophetic teacher with balanced \emph{Propheter} learning.}
\label{alg:one_stage_train}
\begin{algorithmic}[1]
    
\For{$e$ in epochs}
    \If{ $e\ \% \ \text{period} = 0 $}
        \State $\tilde{y}_\theta \leftarrow$ prediction through the GN path 
        \State $\mathcal{L} \leftarrow \ell (y||\tilde{y}_\theta)$
    \Else
        \State $\tilde{y}_\theta \leftarrow$ prediction through the STD path
        \State $\mathcal{L} \leftarrow \ell (y||\tilde{y}_\theta)$
     \EndIf
    \State $\theta \leftarrow \theta - \epsilon \frac{\partial \mathcal{L} }{ \partial \theta}$
\EndFor
\end{algorithmic}
\end{algorithm}

During the training phase, the known class labels of the images are utilized to incorporate the Gaussian distribution $\mathbf{E}_{i}$ into the feature representation of each class. The addition of the Gaussian distribution $\mathbf{E}_{i}$ is performed subsequent to the final convolution layer and prior to the fully connected layer. The mean and variance of each category are utilized as learnable parameters and are incorporated into the model's parameters. This enables the optimization of these parameters through backpropagation, allowing for the model to adjust and improve its predictions over the training set $X$ by minimizing the loss function with respect to the model's parameters $\theta$ and $W$:
\begin{equation}
L( X, Y; \theta, W) \mathrel{:=} \sum_{i=1}^n \ell \left( \sigma(\mathbf{z}_i), y_i\right),
\label{eq:ce_loss}
\end{equation}
where $\mathbf{z}_i = \phi_{\theta,W}(x_i)$ is the output of the model, $\sigma(.)$ is the softmax activation function, and $\ell(.)$ is the loss function, like the cross-entropy loss and others.

However, another challenge arises as the labels of samples are unknown during testing, precluding the addition of distinct Gaussian distributions to individual samples, thereby resulting in an inconsistent training-testing scenario. To overcome this, we employ the technique of knowledge distillation. Typically, knowledge distillation aims to align the logits of the student model with those of the teacher model. In our implementation, we adopt a variation that facilitates information transfer at the feature level.

To clarify, we adopt the model that incorporates Gaussian noise during training as our teacher model. Subsequently, through the utilization of knowledge distillation, we empower the student model to acquire the Gaussian distribution learned by the teacher model in the first stage. The objective loss function described in Eq.~\ref{eq:ce_loss} transforms as follows:
\begin{align}
L( X, Y; \theta, W) \mathrel{:=} \sum_{i=1}^n \ell\left( \sigma(\mathbf{z}_i), y_i\right) + \alpha \cdot \ell_F \left( \mathbf{v}_i, \widehat{\mathbf{v}_i}\right).
\label{eq:featdistill}
\end{align}
Specifically, $\widehat{\mathbf{v}_i}$ represents the feature representation produced by the teacher model, $\mathbf{v}_i$ for the student model, while $\ell_F(\mathbf{v}_i, \widehat{\mathbf{v}_i})$ represents the mean squared error loss that seeks to minimize the dissimilarity between two feature representations.
The hyperparameter $\alpha$ serves as a scaling factor, and unless specified otherwise, is set to 
1.0. Our second stage algorithm is outlined in Alg.~\ref{alg:two_stage_train}.

\begin{algorithm}[t]
\caption{Training a long-tailed student classification model, guided by the learned \emph{Propheter} in Alg.~\ref{alg:one_stage_train}.}
\label{alg:two_stage_train}
\begin{algorithmic}[1]
\For{$e$ in epochs}
    \If{ teacher model }
        \State $\tilde{\widehat{\mathbf{v}}}_i \leftarrow$ feature descriptor produced by teacher model through the GN path
    \ElsIf{student model}
        \State $\tilde{y}_{\theta_{s}} \leftarrow$ prediction through the STD path
        \State $\tilde{{\mathbf{v}}}_i \leftarrow$ feature descriptor produced by student model
        \State $\mathcal{L} \leftarrow \ell (y || \tilde{y}_\theta) + \alpha \cdot \ell_F({\mathbf{v}}_i || \widehat{\mathbf{v}}_i)$
     \EndIf
    \State $\theta_s \leftarrow \theta_s - \epsilon \frac{\partial \mathcal{L} }{ \partial \theta_s}$
\EndFor
\end{algorithmic}
\end{algorithm}

\subsection{Propheter-Guided Long-Tailed Classification }
\label{sect:mainmethod}

With the well-trained Prophetic Teacher (Propheter), the target long tail classification will be trained on the benchmark dataset under Propheter's guidance.

In general, the two-stage approaches involve the utilization of instance sampling in the first stage and class-balanced sampling in the second stage~\cite{chu2020feature,kang2019decoupling}. 
However, it was observed that class-balanced sampling can lead to a reduction in accuracy during the second stage when employing knowledge distillation. As a result, this study only employs instance sampling, 
which not only enhances the efficiency of training but also mitigates the overfitting of the tail classes.

For the second stage, three methods were employed to directly transfer information at the feature level.\\
$\blacktriangleright$ \textbf{1. Classical Decoupling.} The first method resembles the classical decoupling approach, with the learning rate being restarted and trained for a specified number of epochs, followed by the loading of the weights of the teacher model produced in the first stage by the student model.\\
$\blacktriangleright$ \textbf{2. Distillation with High-confidence Kernels.} The second method is similar to the first, but it involves selecting a pre-determined number of high-confidence pictures from the training set, counting their activation values on the last convolution layer, and then selecting the top $k$ (where $k=10$) convolution kernels for each category. During fine-tuning, only the Gaussian distribution is added to the convolution kernels corresponding to these categories on the teacher model. This approach was found to slightly improve the performance of the student model and is considered an extension of the first method.\\
$\blacktriangleright$ \textbf{3. Distilling from Scratch.} The third method involves training the student model from scratch while incorporating a feature distillation loss. Unlike the previous methods, this approach does not employ the use of high\-confidence images to select the convolution kernels of each category, as it is believed that adding this prior knowledge on the teacher model would result in undue prejudice to the student model when training it from scratch. Experiments show that this scheme can maximize the performance of the model.
In the following experimental results, we will only present the results of this method.

\section{Experiments}

\subsection{Datasets and Implementation Details}

\noindent\textbf{Datasets.}
In our experiments, we utilize eight long-tailed benchmark datasets.
\begin{compactitem}
\item\emph{CIFAR-10-LT/CIFAR-100-LT}, are modified versions of CIFAR-10/CIFAR-100, as introduced in \cite{cao2016deep}. They have been generated by downsampling per-class training examples with exponential decay functions, resulting in long-tailed datasets. We perform experiments with imbalance factors of 100, 50, and 10, thereby obtaining a total of six datasets.
\item\emph{ImageNet-LT}, was introduced by Liu \etal~\cite{liu2018open} and is created by artificially truncating the balanced version of ImageNet. It comprises 115.8K images from 1000 categories, with a maximum of 1280 images per class and a minimum of 5 images per class.
\item\emph{Places-LT}, was introduced by Liu \etal~\cite{liu2018open}, is a long-tailed variant of Places365. It comprises 62,500 training images distributed over 365 categories with a severe imbalance factor of 996.
\end{compactitem}

\noindent\textbf{Network Architectures.}
To maintain parity with prior literature, we adopt the network architectures utilized in prior studies~\cite{kang2019decoupling}. Specifically, we utilize the ResNet32 architecture on the CIFAR-10-LT/CIFAR-100-LT datasets, the ResNeXt50 architecture on the ImageNet-LT dataset, and the ResNet152 architecture on the Places-LT datasets.

\noindent\textbf{Evaluation Protocol.}
Our evaluation protocol entails training on the class-imbalanced, long-tailed training set of each dataset and evaluating performance on its corresponding balanced validation or test set. 
The performance metrics are reported on the validation sets.

\noindent\textbf{Parameter Settings.}
The most crucial parameters in the first stage were the parameters for scaling $a_{i}$ and translating $b_{i}$. We adopt a randomized approach to initialize these two parameters and impose constraints on $a_{i}$ to ensure that its value does not fall below 0. In regards to $b_{i}$, we clamp it to the range [0, 1] in order to prevent the occurrence of exploding gradients. 
The cycle period is set to 7. 
More implementation details can be found in the source code.

\subsection{Experimental Results}

\begin{table*}[!t]
\caption{Classification accuracy of ResNet32 trained with different loss functions on long-tailed CIFAR-10 and CIFAR-100.}
\footnotesize
\centering
\resizebox{\textwidth}{!}
{
\input{table/cifar.tex}

}

\label{tab:cifar_results}
\end{table*}

\subsubsection{Results on CIFAR-10-LT/CIFAR-100-LT}
In this study, we conduct extensive evaluations on long-tailed CIFAR datasets with varying degrees of class imbalance. The results, depicted in Tab.~\ref{tab:cifar_results}, showcase the classification accuracy of ResNet-32 on the test set. Our evaluations encompass various loss functions, including Cross-Entropy Loss with softmax activation, Focal Loss, Class-Balanced Loss, and Balanced Softmax Loss, among others. The two entries in the table correspond to two different scenarios. The first shows the baseline performance without any additional modifications, the second corresponds to the result of distillation for 200 epochs (Distillation from scratch). 
All results are averages of three runs.

The results indicate that the addition of a Gaussian distribution layer and the utilization of knowledge distillation enhances the overall performance of the model. Specifically, for the case of Cross-Entropy Loss, the performance improvement surpasses 4\% relative to the baseline. The results of our experiments show that our approach is suitable for long-tailed datasets with various imbalance factors. At the same time, our strategy can increase the model precision of the majority of known loss functions. That demonstrates our generalizability.

\subsubsection{Results on Large-Scale Datasets}

\begin{table}[!t]
\caption{Classification accuracy on the ImageNet-LT and Places-LT dataset. Here, $^\dagger$ indicates that the hyperparameter $\alpha$ is 2.0 for Eq.~\ref{eq:featdistill}.}
\centering
{\input{table/acc_large_dataset}}
\label{tab:large_dataset_results}
\end{table}

We give the results for the baseline and the results following certain distillation epochs (90 epochs for ImageNet-LT and 30 epochs for Places-LT) for the large-scale datasets. For ImageNet-LT and Places-LT, the student model does not load the pre-training weight generated in the first stage (Distillation from scratch). Unfortunately, several methods do not provide hyperparameters for ImageNet-LT and Places-LT, so we only trained the model from scratch using softmax Cross-Entropy Loss, Focal Loss, and Balanced Softmax Loss. 
The results are shown in Tab.~\ref{tab:large_dataset_results}.

The experimental results show that our method still improves the performance of the model on large-scale datasets. This demonstrates that our strategy is still useful for large datasets and that it can enhance the performance of the model based on existing methods.

\subsection{Ablation Study}
We investigate the effect of (1) the proposed three methods by us, (2) the number and placement of Gaussian distribution layers. 
Our study focuses on the CIFAR-10-LT dataset with an imbalance factor of 100 and the use of the Balanced Softmax Loss.

\begin{table}[t]
\caption{Overall classification accuracies of {various knowledge transfer methods} elaborated in Sect.~\ref{sect:mainmethod}.}
\small
\footnotesize
\centering
{\renewcommand{\arraystretch}{1.2}
\begin{tabular}{ lccc }
\toprule
\multirow{1}{*}{\bf \#}
& \multicolumn{1}{l}{\bf Methods}
& \multicolumn{1}{c}{\bf Accuracy}
& \multicolumn{1}{c}{\bf Gains}
\\
\hline

1 & \multicolumn{1}{l}{Baseline}
& \multicolumn{1}{c}{81.91}
& \multicolumn{1}{c}{-}
\\

2 & \multicolumn{1}{l}{Classical Decouple}
& \multicolumn{1}{c}{82.63}
& \multicolumn{1}{c}{\bf 0.72 $\uparrow$}
\\

3 & \multicolumn{1}{l}{Distillation with High-confidence Kernels}
& \multicolumn{1}{c}{82.81}
& \multicolumn{1}{c}{\bf 0.90 $\uparrow$}
\\

4 & \multicolumn{1}{l}{Distilling from Scratch}
& \multicolumn{1}{c}{83.99}
& \multicolumn{1}{c}{\bf 2.08 $\uparrow$}
\\

\bottomrule

\end{tabular}}

\label{tab:different_method_result}
\vspace{-2mm}
\end{table}

\noindent\textbf{Various Knowledge Transfer Methods in Sect.~\ref{sect:mainmethod}.}
We contrast the outcomes of the three suggested approaches with the baseline, as indicated in Tab.~\ref{tab:different_method_result}.
 We can find that the three methods we proposed have improved the performance of the model, but the performance of method 3 (Distillation from scratch) has improved the most. At the same time, by comparing the second and third rows of the table, if we add Gaussian distribution on the class-related convolution kernel in the classical distillation scheme, we can slightly improve the performance of the model. This also proves the effectiveness of method 2 (Distillation with high-confidence kernels) from the side.

\noindent\textbf{Position and Number of Gaussian Distribution Layers.} In the ResNet32 architecture, there are three blocks. To assess the impact of the number and position of Gaussian distribution layers, we conduct experiments by altering their placement. 
The results are shown in Tab.~\ref{tab:abs_gn}.
The first row serves as the baseline for comparison.

\begin{table}[t]
\caption{The influence of the position and number of Gaussian distribution layers on performance. Here, $\checkmark$ means that a Gaussian distribution layer is added after the corresponding block.}
\begin{center}
    \begin{tabular}{ccccc}
    \toprule
    \multirow{1}{*}{\#} & Block1 & \multicolumn{1}{c}{Block2} & \multicolumn{1}{c}{Block3} & Accuracy (\%) \\
    \midrule
    1 & - & - & $\checkmark$ & 83.99 \\ 
    2 & - & $\checkmark$ & $\checkmark$ & 81.11 \\ 
    3 & $\checkmark$ & $\checkmark$ & $\checkmark$ & 80.40 \\ 
    4 & $\checkmark$ & - & - & 80.48 \\ 
    5 & - & $\checkmark$ & - & 80.87 \\ 
	\bottomrule
    \end{tabular}
\end{center}

\label{tab:abs_gn}
\vspace{-4mm}
\end{table}

Our findings indicate that adding a single Gaussian distribution layer after the third block resulted in the most favorable outcome. This is evident when comparing the table's first, second, and third rows. Additionally, the comparison between the first, fourth, and fifth rows highlights that placing a single Gaussian distribution layer after the third block appears to be the optimal choice. We believe that this is because the features generated by the last block are more strongly associated with the classes, 
thereby leading to optimal performance.

\section{Conclusions}
In this work, we present a novel two-stage long-tailed recognition strategy that incorporates instance balance sampling in both stages and utilizes a singular teacher model in the process of knowledge distillation.
Our method essentially learns the residual between the distribution learned by the model for each class and the true distribution in the latent space, from the perspective of the latent space.
Our proposed method has been demonstrated to enhance the performance of existing long-tailed classification techniques and achieve substantial gains, as validated through a comprehensive series of experiments. In our future work, we will strive to explore the extension of our method to transformer-based models, and even the fields of object detection and image segmentation.

\subsubsection{Acknowledgements}
This work is funded by National Key Research and Development Project (Grant No: 2022YFB2703100), National Natural Science Foundation of China (61976186, U20B2066), Zhejiang Province High-Level Talents Special Support Program ``Leading Talent of Technological Innovation of Ten-Thousands Talents Program'' (No. 2022R52046), Ningbo Natural Science Foundation (2022J182), Basic Public Welfare Research Project of Zhejiang Province (LGF21F020020), and the Fundamental Research Funds for the Central Universities (2021FZZX001-23, 226-2023-00048).
This work is partially supported by the National Natural Science Foundation of China (Grant No. 62106235), the Exploratory Research Project of Zhejiang Lab (2022PG0AN01), and the Zhejiang Provincial Natural Science Foundation of China (LQ21F020003).

%
%
%
\bibliographystyle{splncs04}
\bibliography{egbib}
\end{document}

%% file: table/cifar.tex
\begin{tabular}{ lcccccc }
\hline
\multicolumn{1}{l}{Dataset Name}
& \multicolumn{3}{|c}{Long-Tailed CIFAR-10}
& \multicolumn{3}{|c}{Long-Tailed CIFAR-100}
\\ \hline

\multicolumn{1}{l}{Imbalance} 
& \multicolumn{1}{|c}{100}
& \multicolumn{1}{|c}{50}
& \multicolumn{1}{|c}{10}
& \multicolumn{1}{|c}{100}
& \multicolumn{1}{|c}{50}
& \multicolumn{1}{|c}{10}
\\ \hline

\multicolumn{1}{l}{CE Loss} 
& \multicolumn{1}{|l}{75.24}
& \multicolumn{1}{|l}{79.73 }
& \multicolumn{1}{|l}{88.90 }
& \multicolumn{1}{|l}{40.29 }
& \multicolumn{1}{|l}{45.06 }
& \multicolumn{1}{|l}{58.69 }
\\ \hdashline

\multicolumn{1}{l}{+Ours} 
& \multicolumn{1}{|l}{ 79.27 (\textcolor{red}{+4.03})}
& \multicolumn{1}{|l}{ 83.34 (\textcolor{red}{+3.61})} 
& \multicolumn{1}{|l}{ 89.52 (\textcolor{red}{+0.62})} 
& \multicolumn{1}{|l}{ 42.07 (\textcolor{red}{+1.78})} 
& \multicolumn{1}{|l}{ 47.12 (\textcolor{red}{+2.06})} 
& \multicolumn{1}{|l}{ 60.14 (\textcolor{red}{+1.45})} 
\\ \hline

\multicolumn{1}{l}{Focal Loss ($\gamma = 1.0$)~\cite{Focalloss}}
& \multicolumn{1}{|l}{75.61 } 
& \multicolumn{1}{|l}{79.17 } 
& \multicolumn{1}{|l}{87.90 } 
& \multicolumn{1}{|l}{40.69 } 
& \multicolumn{1}{|l}{45.57 } 
& \multicolumn{1}{|l}{58.06 }
\\ \hdashline

\multicolumn{1}{l}{+Ours}
& \multicolumn{1}{|l}{ 77.12 (\textcolor{red}{+1.51})}  
& \multicolumn{1}{|l}{ 81.73 (\textcolor{red}{+2.56})}  
& \multicolumn{1}{|l}{ 89.05 (\textcolor{red}{+1.15})}  
& \multicolumn{1}{|l}{ 41.94 (\textcolor{red}{+1.25})}  
& \multicolumn{1}{|l}{ 46.83 (\textcolor{red}{+1.26})}  
& \multicolumn{1}{|l}{ 60.02 (\textcolor{red}{+1.96})}  
\\ \hline

\multicolumn{1}{l}{Focal Loss ($\gamma = 2.0$)~\cite{Focalloss}}
& \multicolumn{1}{|l}{74.26 } 
& \multicolumn{1}{|l}{78.25 } 
& \multicolumn{1}{|l}{87.31 } 
& \multicolumn{1}{|l}{40.22 } 
& \multicolumn{1}{|l}{45.90 } 
& \multicolumn{1}{|l}{57.08 } 
\\ \hdashline

\multicolumn{1}{l}{+Ours}
& \multicolumn{1}{|l}{ 76.57 (\textcolor{red}{+2.31})}  
& \multicolumn{1}{|l}{ 81.12 (\textcolor{red}{+2.87})}  
& \multicolumn{1}{|l}{ 88.54 (\textcolor{red}{+1.23})} 
& \multicolumn{1}{|l}{ 41.50 (\textcolor{red}{+1.28})}  
& \multicolumn{1}{|l}{ 46.72 (\textcolor{red}{+0.82})}  
& \multicolumn{1}{|l}{ 59.75 (\textcolor{red}{+2.67})}  
\\ \hline

\multicolumn{1}{l}{Class-Balanced Loss~\cite{cui2019CBLoss}}
& \multicolumn{1}{|l}{73.48 } 
& \multicolumn{1}{|l}{80.60 } 
& \multicolumn{1}{|l}{87.66 } 
& \multicolumn{1}{|l}{39.79 } 
& \multicolumn{1}{|l}{45.89 } 
& \multicolumn{1}{|l}{58.97 } 
\\ \hdashline

\multicolumn{1}{l}{+Ours}
& \multicolumn{1}{|l}{ 76.71 (\textcolor{red}{+3.23})}  
& \multicolumn{1}{|l}{ 81.43 (\textcolor{red}{+0.83})} 
& \multicolumn{1}{|l}{ 88.45 (\textcolor{red}{+0.79})} 
& \multicolumn{1}{|l}{ 41.77 (\textcolor{red}{+1.98})} 
& \multicolumn{1}{|l}{ 48.91 (\textcolor{red}{+3.02})} 
& \multicolumn{1}{|l}{ 59.94 (\textcolor{red}{+0.97})} 
\\ \hline

\multicolumn{1}{l}{Balanced Softmax Loss~\cite{Ren2020balms}}
& \multicolumn{1}{|l}{81.91 } 
& \multicolumn{1}{|l}{84.27 } 
& \multicolumn{1}{|l}{89.92 } 
& \multicolumn{1}{|l}{46.93 }
& \multicolumn{1}{|l}{51.24 } 
& \multicolumn{1}{|l}{60.66 } 
\\ \hdashline

\multicolumn{1}{l}{+Ours}
& \multicolumn{1}{|l}{ 83.99 (\textcolor{red}{+2.08})} 
& \multicolumn{1}{|l}{ 86.05 (\textcolor{red}{+1.78})} 
& \multicolumn{1}{|l}{ 90.23 (\textcolor{red}{+0.31})} 
& \multicolumn{1}{|l}{ 48.33 (\textcolor{red}{+1.40})}
& \multicolumn{1}{|l}{ 52.76 (\textcolor{red}{+1.52})}
& \multicolumn{1}{|l}{ 62.23 (\textcolor{red}{+1.57})}
\\ \hline

\end{tabular}

%% file: table/acc_large_dataset.tex
\begin{tabular}{ lcccccc }
\toprule

\multicolumn{1}{l}{}
& \multicolumn{3}{c}{\bf ImageNet-LT}
& \multicolumn{3}{c}{\bf Places-LT}
\\
\cmidrule{2-4}
\cmidrule{5-7}

\multicolumn{1}{l}{}
& \multicolumn{1}{c}{Baseline}
& \multicolumn{1}{c}{Ours}
& \multicolumn{1}{c}{Gains}

& \multicolumn{1}{c}{Baseline}
& \multicolumn{1}{c}{Ours}
& \multicolumn{1}{c}{Gains}
\\ \midrule

\multicolumn{1}{l}{CE Loss} 
& \multicolumn{1}{c}{43.97 }
& \multicolumn{1}{c}{44.57 }
& \multicolumn{1}{c}{\bf 0.60 $\uparrow$} 

& \multicolumn{1}{c}{ 29.32 }
& \multicolumn{1}{c}{ 30.11 } 
& \multicolumn{1}{c}{\bf 0.79 $\uparrow$ } 
\\ 

\multicolumn{1}{l}{Focal Loss ($\gamma = 1.0$)~\cite{Focalloss}}
& \multicolumn{1}{c}{ 43.47 }
& \multicolumn{1}{c}{ 43.76 } 
& \multicolumn{1}{c}{\bf 0.29 $\uparrow$} 

& \multicolumn{1}{c}{ 29.28 }
& \multicolumn{1}{c}{ 30.12 } 
& \multicolumn{1}{c}{\bf 0.84 $\uparrow$}  
\\ 

\multicolumn{1}{l}{Focal Loss ($\gamma = 2.0$)~\cite{Focalloss}}
& \multicolumn{1}{c}{ 43.29 }
& \multicolumn{1}{c}{ 44.26$^\dagger$ } 
& \multicolumn{1}{c}{\bf 0.97 $\uparrow$} 

& \multicolumn{1}{c}{ 29.18 }
& \multicolumn{1}{c}{ 29.78 } 
& \multicolumn{1}{c}{\bf 0.60 $\uparrow$} 
\\

\multicolumn{1}{l}{Balanced Softmax Loss~\cite{Ren2020balms}}
& \multicolumn{1}{c}{ 47.95 }
& \multicolumn{1}{c}{ 48.55 } 
& \multicolumn{1}{c}{\bf 0.60 $\uparrow$} 

& \multicolumn{1}{c}{ 35.60 }
& \multicolumn{1}{c}{ 36.55 } 
& \multicolumn{1}{c}{\bf 0.95 $\uparrow$} 
\\ 

\bottomrule

\end{tabular}